\documentclass[journal]{IEEEtran}
\usepackage{color}

\usepackage{graphicx}
\usepackage[numbers,sort&compress]{}
\usepackage{textcomp}
\usepackage[tight]{subfigure}

\usepackage{cite}

\ifCLASSINFOpdf
\else
\fi
\usepackage{amsmath}
\usepackage{url}

\begin{document}
%
\title{Scale-Space Anisotropic Total Variation for Limited Angle Tomography}
%
%
%

\author{Yixing~Huang,
        Oliver~Taubmann,
        Xiaolin~Huang,
        Viktor~Haase,
        Guenter~Lauritsch,
        and~Andreas~Maier
\thanks{Y.~Huang is with Pattern Recognition Lab, Friedrich-Alexander-University Erlangen-Nuremberg, Erlangen, Germany (e-mail: yixing.yh.huang@fau.de).}
\thanks{O.~Taubmann, and A.~Maier are with Pattern Recognition Lab, Friedrich-Alexander-University Erlangen-Nuremberg, Erlangen, Germany, and also with Erlangen Graduate School in Advanced Optical Technologies (SAOT), Erlangen, Germany.}
\thanks{X.~Huang was with Pattern Recognition Lab, Friedrich-Alexander-University Erlangen-Nuremberg, Erlangen, Germany and now is with Institute of Image Processing and Pattern Recognition, Shanghai Jiao Tong University, Shanghai, China.}
\thanks{V.~Haase is with Pattern Recognition Lab, Friedrich-Alexander-University
Erlangen-Nuremberg, Erlangen, Germany, and with Siemens Healthcare
GmbH, Forchheim, Germany, and also with the Department of Radiology, University of Utah,
Salt Lake City, Utah, USA.
}
\thanks{G.~Lauritsch and O.~Taubmann are with Siemens Healthcare GmbH, Forchheim, Germany.}
}

%
%

\markboth{Journal of \LaTeX\ Class Files,~Vol.~14, No.~8, August~2017}%
{Shell \MakeLowercase{\textit{et al.}}: Bare Demo of IEEEtran.cls for IEEE Journals}
%



\maketitle
\begin{abstract}
This paper addresses streak reduction in limited angle tomography. Although the iterative reweighted total variation (wTV) algorithm reduces small streaks well, it is rather inept at eliminating large ones since total variation (TV) regularization is scale-dependent and may regard these streaks as homogeneous areas. Hence, the main purpose of this paper is to reduce streak artifacts at various scales. We propose the scale-space anisotropic total variation (ssaTV) algorithm in two different implementations. The first implementation (ssaTV-1) utilizes an anisotropic gradient-like operator which uses $2 \cdot s$ neighboring pixels along the streaks' normal direction at each scale $s$. The second implementation (ssaTV-2) makes use of anisotropic down-sampling and up-sampling operations, similarly oriented along the streaks' normal direction, to apply TV regularization at various scales. Experiments on numerical and clinical data demonstrate that both ssaTV algorithms reduce streak artifacts more effectively and efficiently than wTV, particularly when using multiple scales.
\end{abstract}

\begin{IEEEkeywords}
limited angle tomography, streak artifacts, total variation, anisotropic, scale-space.
\end{IEEEkeywords}

%
\IEEEpeerreviewmaketitle

\section{Introduction}
\begin{figure}[t]
\centering
\begin{minipage}[t]{0.3\linewidth}
\subfigure[Custom phantom]{
\includegraphics[width=1\textwidth]{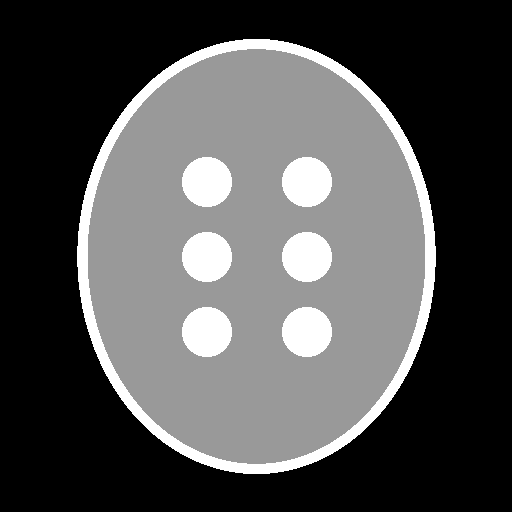}
}
\end{minipage}
\begin{minipage}[t]{0.3\linewidth}
\subfigure[FBP reconstruction]{
\includegraphics[width=1\textwidth]{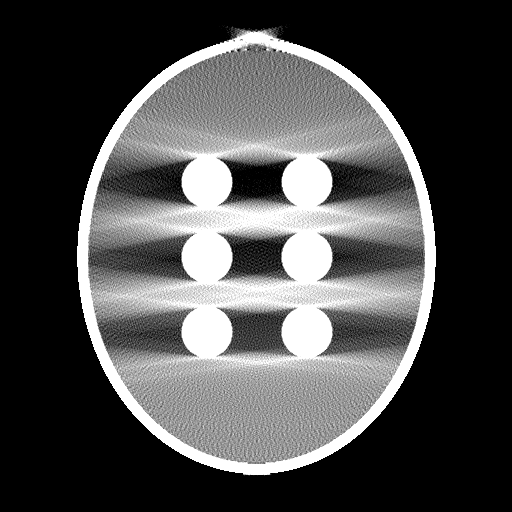}
\label{subfig:artifactsDemonSubfig}
}
\end{minipage}
\begin{minipage}[t]{0.37\linewidth}
\subfigure[Limited angle scan]{
\includegraphics[width=1\textwidth]{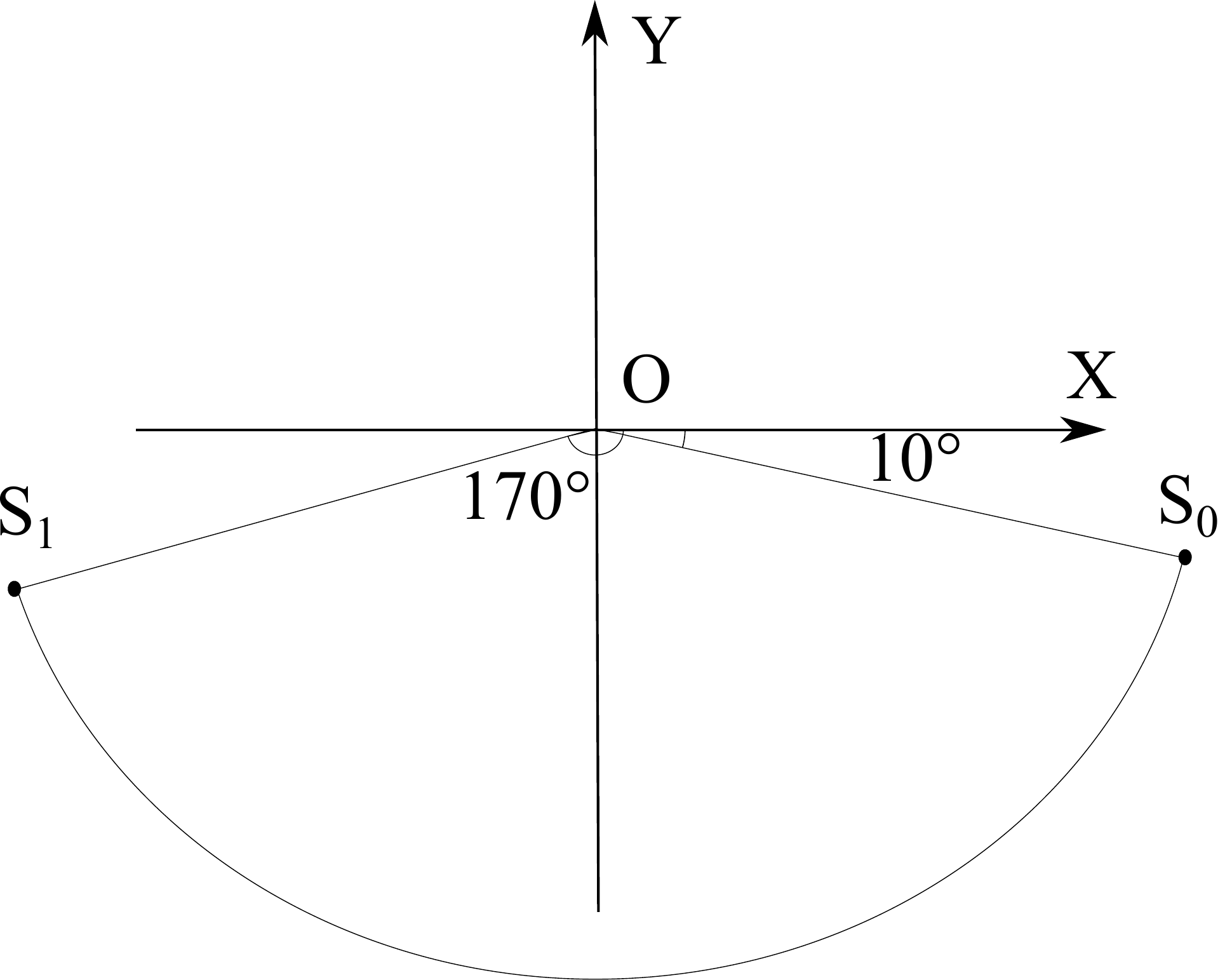}
\label{fig:trajectory}
}
\end{minipage}
\caption{Demonstration of artifacts in limited angle tomography: a custom phantom (a) and its filtered back-projection (FBP) reconstruction (b) from $160^\circ$ limited angle sinogram acquired in a limited angle scan (c) in fan-beam geometry, window: [-300, 200]\,HU.}
\label{fig:artifactsDemon}
\end{figure}

%
%
%
%
\IEEEPARstart{C}{one-beam} computed tomography (CBCT) is a widely used medical imaging technology. CBCT reconstructs a volume of data which provides information about the anatomical morphology of the patient. In order to get a complete set of projection data for reconstruction, currently most CBCT systems need the X-ray source and detector to rotate around 200$^\circ$, which is called a short scan. In practical applications, CBCT systems, particularly angiographic C-arm devices, are used to acquire 3-D images for planning, guiding, and monitoring of interventional operations. In these situations, the gantry rotation might be restricted by other system parts or external obstacles. In this case, only limited angle data are acquired. Image reconstruction from data acquired in an insufficient angular range is called limited angle tomography. Due to data insufficiency, artifacts, typically in the form of streaks, will occur in the reconstructed images (Fig.~\ref{subfig:artifactsDemonSubfig}). They cause boundary distortion, intensity leakage, and edge blurry. The characterization of streak artifacts can be found in \cite{frikel2013characterization,nguyen2015strong}. Generally, streak artifacts appear at boundary areas, which is object dependent. However, the orientations of streak artifacts are highly dependent on the scan trajectory, mostly at the missing angular ranges. For example, when the scan is from $10^\circ-170^\circ$ shown in Fig.~\ref{fig:artifactsDemon}, most streaks are approximately oriented at the horizontal direction, especially for the low frequency streaks (large inhomogeneities) between circular areas. These streak artifacts degrade the image quality and may lead to misinterpretation of the images. Therefore, streak reduction in limited angle tomography has important clinical value.

A lot of effort has already gone into suppressing streak artifacts in limited angle tomography \cite{natterer2001mathematics,quinto2006introduction}. One approach is to extrapolate/interpolate the missing data in projection domain \cite{louis1980picture,willsky1990constrained,qu2008iterative,huang2017Restoration}. Another approach is compressed sensing, which has attracted tremendous attention since it requires only relatively little data to obtain a good reconstruction result by exploiting sparsity \cite{candes2006robust, sidky2008image}. In particular, iterative reconstruction algorithms regularized by total variation (TV) were demonstrated to be effective in streak reduction in limited angle tomography \cite{sidky2006accurate, iTV}.

However, images processed by total-variation-based methods typically exhibit an undesirable staircasing effect which transforms gray value slopes into stair-like shapes and causes edges to be blurred and fine structures to be removed \cite{Staircasing, jalalzai2016some}. In 2008, Cand\`es et al.~\cite{Candes} proposed the iterative reweighted TV (wTV) algorithm to enhance sparsity in the gradient domain more effectively, which reduces the staircasing effect intrinsically. The wTV algorithm has been widely applied to different fields of image processing and its advantages are well-understood \cite{needell2009noisy, zhengguang2013efficient, shi2014adaptive, kohler2016robust}. Therefore, in this paper, we adapt wTV for use in limited angle reconstruction. Problems optimized with wTV are generally nonconvex \cite{ochs2015iteratively}. The convergence of wTV algorithms has been proved by \cite{chen2014convergence,ochs2015iteratively}. However, the converged stationary point might still be a local minimum instead of the global one \cite{chen2014convergence,ochs2015iteratively}. 
In limited angle tomography, the reconstruction problem is severely ill-posed \cite{davison1983ill}. Therefore, very likely only a local optimum is obtained. TV regularization generally is scale-dependent~\cite{DavidStrong}. It typically uses two neighboring pixels to compute the derivative in each direction and the resulting gradient operators are hardly able to detect variations on larger scales. Specifically, in limited angle tomography large streaks with low frequency and a high intensity difference may be regarded as proper structures by the wTV algorithm.

In many imaging processing tasks, scale-space optimization approaches are widely used to avoid local minima and accelerate the convergence speed \cite{mjolsness1991multiscale,lucia2004multi,hager2006multiscale}. To jointly reduce streaks of various scales more effectively, we perform the regularization in scale space since each individual scale is most sensitive to artifacts of a specific spatial extent \cite{adelson1984pyramid, chambolle2001interpreting, lindeberg2013scale}. 

As mentioned above, in limited angle tomography, shape and orientation of streak artifacts are closely related to the missing angular range in the acquisition. Making use of such prior knowledge, anisotropic TV (aTV) methods are designed \cite{moll2005anisotropic, grasmair2010anisotropic,chen2013limited,huang2016weighted,huang2016New,wang2017reweighted}. For instance, Chen et al.~\cite{chen2013limited} assigned different weighting factors to different directions, which resulted in better performance on edge recovery and streak artifact reduction than isotropic TV. Wang et al.~\cite{wang2017reweighted} proposed to combine wTV with another form of aTV to prevent blurring of certain orientation edges. This approach works well on simple tubular structures, but appears to struggle with more complex objects.
In our paper, we incorporate anisotropy into the wTV algorithm at each scale and propose two different aTV implementations. The first one utilizes an anisotropic gradient-like operator which uses $2 \cdot s$ neighboring pixels at each scale $s$ to calculate the gradient along the streaks' normal direction \cite{huang2016weighted}. The second one makes use of anisotropic down-sampling and up-sampling operations, similarly oriented along the streaks' normal direction \cite{huang2016New}. To validate the advantages of our proposed scale-space anisotropic total variation (ssaTV) algorithms, experiments on both numerical and clinical data are performed.

\section{Materials and methods}
\subsection{Baseline: Iterative reweighted total variation (wTV)}
A scheme combining the simultaneous algebraic reconstruction technique (SART) with TV minimization \cite{sidky2006accurate,sidky2008image, iTV, chen2013limited}, namely SART-TV, is often utilized for TV-regularized iterative reconstructions. It alternatively minimizes a data fidelity term $||\boldsymbol{A}\boldsymbol{f}-\boldsymbol{p}||_2^2$ and a TV term $||\boldsymbol{f}||_{\text{TV}}$,
where $\boldsymbol{f}$ consists of the voxels of the desired image stacked in a column vector, $\boldsymbol{A}$ is the system matrix and $\boldsymbol{p}$ is the vector of acquired projection data. The data fidelity term is optimized by the following SART update \cite{andersen1984simultaneous},
\begin{equation}
\boldsymbol{f}_j^{n+1}=\boldsymbol{f}_j^{n}+\lambda \cdot \frac{\sum_{\boldsymbol{p}_i\in \boldsymbol{p}_{\beta}}\frac{\boldsymbol{p}_i-\sum^N_{k=1}A_{i,k}\cdot \boldsymbol{f}_k^n}{\sum_{k=1}^{N}A_{i,k}}\cdot A_{i,j}}{\sum_{\boldsymbol{p}_i\in \boldsymbol{p}_{\beta}}A_{i,j}},
\label{eqn:SART}
\end{equation}
where $j$ is the pixel index of $\boldsymbol{f}$, $i$ is the projection ray index of $\boldsymbol{p}$, $A_{i,j}$ is the element of $A$ at the $i$-th row and the $j$-th column, $n$ is the iteration number, $\beta$ is the X-ray source rotation angle, $N$ is the total number of pixels in $\boldsymbol{f}$, $\lambda$ is a relaxation parameter, and $\boldsymbol{p}_i\in \boldsymbol{p}_{\beta}$ stands for the projection rays when the X-ray source is at position $\beta$.

We define the regular non-weighted TV term $||\boldsymbol{f}||_{\text{TV}}$ as,
\begin{equation}
||\boldsymbol{f}||_{\text{TV}}=\sum_{x,y,z}||\mathcal{D}\boldsymbol{f}_{x,y,z}||,
\end{equation} 
where $x$, $y$ and $z$ are the spatial indices of a voxel into the 3-D grid, $||\cdot||$ is the Euclidean norm and $\mathcal{D}$ is a conventional isotropic gradient operator,
\begin{equation}
\mathcal{D}\boldsymbol{f}_{x,y,z}=\left( \mathcal{D}_x\boldsymbol{f}_{x,y,z}, \mathcal{D}_y\boldsymbol{f}_{x,y,z}, \mathcal{D}_z\boldsymbol{f}_{x,y,z}\right),
\end{equation}
with $\mathcal{D}_x$, $\mathcal{D}_y$ and $\mathcal{D}_z$ being discrete derivative operators along the coordinate axes, 
\begin{equation}
\begin{split}
&\mathcal{D}_x\boldsymbol{f}_{x,y,z}=\boldsymbol{f}_{x,y,z}-\boldsymbol{f}_{x-1,y,z},\\
&\mathcal{D}_y\boldsymbol{f}_{x,y,z}=\boldsymbol{f}_{x,y,z}-\boldsymbol{f}_{x,y-1,z},\\
&\mathcal{D}_z\boldsymbol{f}_{x,y,z}=\boldsymbol{f}_{x,y,z}-\boldsymbol{f}_{x,y,z-1}.
\end{split}
\label{eqn:gradientOperator}
\end{equation}

For the wTV algorithm, the TV term $||\boldsymbol{f}||_{\text{TV}}$ is extended to $||\boldsymbol{f}||_{\text{wTV}}$ by adding a weighting vector. According to Cand\`es et al.~\cite{Candes},
\begin{equation}
\begin{split}
&||\boldsymbol{f}^{(n)}||_{\text{wTV}}=\sum_{x,y,z}\boldsymbol{w}^{(n)}_{x,y,z}||\mathcal{D}\boldsymbol{f}^{(n)}_{x,y,z}||,\\
&\boldsymbol{w}^{(n)}_{x,y,z}=\frac{1}{||\mathcal{D}\boldsymbol{f}^{(n-1)}_{x,y,z}||+\epsilon},
\end{split}
\label{eq:WeightsUpdate}
\end{equation}
where $\boldsymbol{f}^{(n)}$ is the image at the $n$-th iteration, $\boldsymbol{w}^{(n)}$ is the weight vector for the $n$-th iteration which is computed from the previous iteration, and $\epsilon$ is a small positive number added to avoid division by zero. Cand\`es et al.~recommend that $\epsilon$ should be slightly smaller than the expected nonzero magnitude of $\mathcal{D}\boldsymbol{f}_{x,y,z}$. For simplicity, the iteration index $n$ is kept for the weight vector only and omitted for other variables. 

\begin{figure*}[t]
\centering
\begin{minipage}[b]{0.55\linewidth}
\centering
  \includegraphics[width=\textwidth]{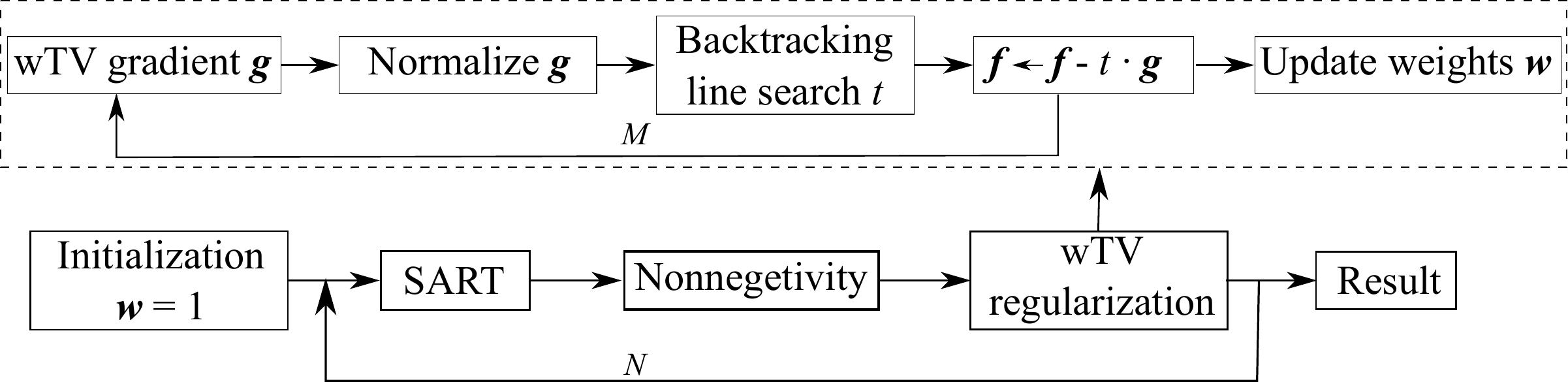}
\end{minipage}
\caption{Our wTV algorithm iterates SART updates and wTV regularization steps alternatively $N$ times in the main loop and repeats the gradient descent process $M$ times in the inner loop.}
  \label{fig:wTValgorithm}
\end{figure*}
A flow chart summarizing our implementation of wTV for limited angle reconstruction is shown in Fig.~\ref{fig:wTValgorithm}. It utilizes the SART-TV scheme similar to \cite{sidky2006accurate}. The main loop iterates at most $N$ times. Each iteration contains a SART update followed by a nonnegativity constraint \cite{sidky2008image} to increase data fidelity and a wTV regularization step using a gradient decent method to minimize the wTV term of the current image. The regularization step, shown within a dashed outline, is separated from the SART update so that it can be replaced by other TV regularization variants in the next sections. The partial derivative of $||\boldsymbol{f}^{(n)}||_{\text{wTV}}$ w.\,r.\,t.~each image voxel is denoted by $\mathbf{g}$,
\begin{equation}
\centering
\begin{split}
&\mathbf{g}_{x,y,z}=\frac{\partial{||\boldsymbol{f}||_{\text{wTV}}}}{\partial {\boldsymbol{f}_{x,y,z}}}\\
&=\boldsymbol{w}^{(n)}_{x,y,z}\cdot \frac{\mathcal{D}_x\boldsymbol{f}_{x,y,z} + \mathcal{D}_y\boldsymbol{f}_{x,y,z}+\mathcal{D}_z\boldsymbol{f}_{x,y,z}}{||\mathcal{D}\boldsymbol{f}_{x,y,z}||}\\
&-\boldsymbol{w}^{(n)}_{x+1,y,z}\cdot \frac{\mathcal{D}_x\boldsymbol{f}_{x+1,y,z}}{||\mathcal{D}\boldsymbol{f}_{x+1,y,z}||}
-\boldsymbol{w}^{(n)}_{x,y+1,z}\cdot \frac{\mathcal{D}_y\boldsymbol{f}_{x,y+1,z}}{||\mathcal{D}\boldsymbol{f}_{x,y+1,z}||}\\
&-\boldsymbol{w}^{(n)}_{x,y,z+1}\cdot \frac{\mathcal{D}_z\boldsymbol{f}_{x,y,z+1}}{||\mathcal{D}\boldsymbol{f}_{x,y,z+1}||},
\end{split}
\label{eq:TVgradient}
\end{equation}
where $\boldsymbol{w}^{(n)}$ is kept as constant at the $n$-th iteration. The partial derivative $\mathbf{g}$ is needed to perform a gradient descent iteration.  $\mathbf{g}$ is normalized to get the direction for gradient descent. A backtracking line search \cite{boyd2004convex} is applied to get the step size $t$. Afterwards the image is updated as $\boldsymbol{f} \gets \boldsymbol{f}-t\cdot\mathbf{g}$. This gradient descent process is repeated $M$ times. After that, $\boldsymbol{w}^{(n)}$ is updated.

\subsection{Scale-space anisotropic total variation (ssaTV)}
The effect of conventional TV regularization, including wTV, is limited in spatial scale since only two neighboring pixels are used to compute the derivative in each direction and the resulting gradient operators are hardly able to detect variations on larger scales. This is why small streaks are reduced effectively by wTV while large streaks remain. To reduce large streaks effectively and efficiently, we apply wTV regularization at various spatial scales along the streaks' normal direction using a scale-space approach. For this purpose, two scale-space anisotropic total variation (ssaTV) algorithms are proposed. 

\subsubsection{The first ssaTV algorithm (ssaTV-1)}
 \begin{figure*}[t]
 \centering
\begin{minipage}[b]{0.55\linewidth}
\centering
  \includegraphics[width=\textwidth]{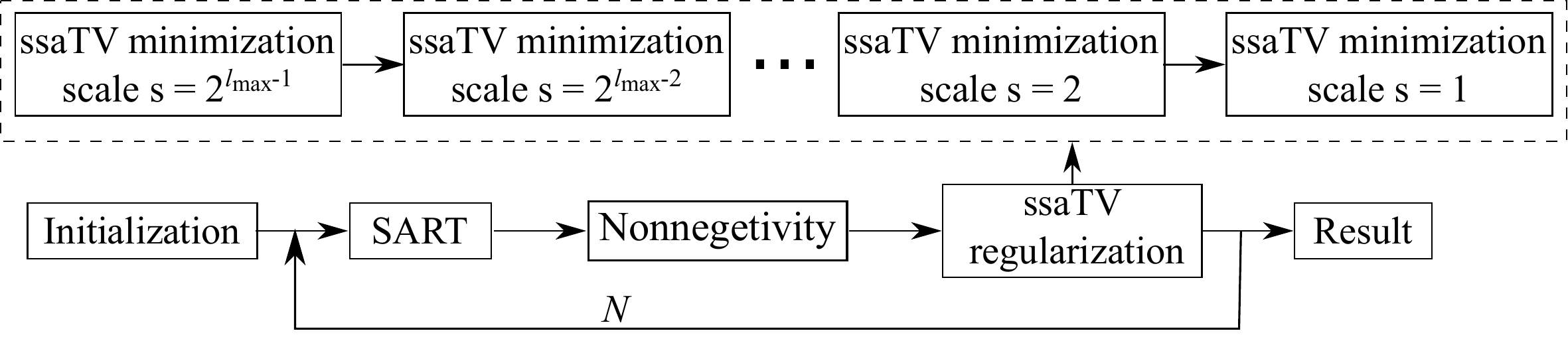}
\end{minipage}
  \caption{The ssaTV algorithm uses multiple scales during regularization. For ssaTV-1, the ssaTV minimization box is the same as the wTV regularization box in Fig.~\ref{fig:wTValgorithm} but uses the modified derivative-like operator $\mathcal{\tilde{D}}_y\boldsymbol{f}$.}
  \label{fig:ssaTValgorithm}
\end{figure*}
In contrast to wTV, ssaTV-1 employs an anisotropic gradient operator. Instead of using two neighboring pixels to define the derivative in each direction, we propose to use more pixels along the normal direction of most streaks. Specifically, $2\cdot s$ neighboring pixels are used at a scale $s$. In practice, streaks can be aligned with a coordinate axis, e.\,g.~the $X$-axis, if we define a coordinate system with the $Y$-axis as the symmetry axis of the scan angular range, e.\,g.~$10^\circ$ - $170^\circ$ (Fig.~\ref{fig:trajectory}). In the given example, most streaks occur in the horizontal direction, which causes more variations along $Y$ than along $X$. Therefore, we introduce a modified derivative-like operator along $Y$ direction, denoted by $\mathcal{\tilde{D}}_y$, to enhance TV regularization at this direction. $\mathcal{\tilde{D}}_y\boldsymbol{f}$ is generally represented as,
\begin{equation}
\mathcal{\tilde{D}}_y\boldsymbol{f}_{x,y,z}=\sum_{i=-s}^{s-1}\boldsymbol{a}_{s+i} \cdot \boldsymbol{f}_{x,y+i,z},
\label{eqn:waTV}
\end{equation}
where $\boldsymbol{a}$ is a derivative-like kernel with $2\cdot s$ elements. This results in an anisotropic gradient operator $\mathcal{\tilde{D}}$ at scale $s$, denoted by $\mathcal{\tilde{D}}$,
\begin{equation}
\mathcal{\tilde{D}}\boldsymbol{f}_{x,y,z}=\left( \mathcal{D}_x\boldsymbol{f}_{x,y,z},\mathcal{\tilde{D}}_y\boldsymbol{f}_{x,y,z}, \mathcal{D}_z\boldsymbol{f}_{x,y,z}\right),
\end{equation}
where $\mathcal{D}_x\boldsymbol{f}$ and $\mathcal{D}_z\boldsymbol{f}$ are identical to those in Eqn.~(\ref{eqn:gradientOperator}). For ssaTV-1, the isotropic gradient operator $\mathcal{D}$ is replaced by the anisotropic one $\mathcal{\tilde{D}}$ in Eqn.~\ref{eq:WeightsUpdate}.

The next step is to design the kernel $\boldsymbol{a}$. The purpose of $\tilde{D}_y$ is to detect variations at coarse scales. Coarse scale structures are typically attained by low-pass filtering,
\begin{equation}
\boldsymbol{f'}_{x, y, z} = \sum_{j = -L}^{L}{\boldsymbol{h}_{L-j}\cdot\boldsymbol{f}_{x, y + j, z}},
\label{eqn:lowPassFiltering}
\end{equation}
where $\boldsymbol{h}$ is a 1-D low-pass kernel with length $2\cdot L+1$ and
$\boldsymbol{f'}$ is the smoothed image computed as the convolution of $\boldsymbol{f}$ and $\boldsymbol{h}$ along $Y$ direction. The Gaussian kernel is the most widely used low-pass filter for scale-space approaches since it satisfies adequate scale-space conditions \cite{babaud1986uniqueness,lindeberg1990scale}. Therefore, we choose the Gaussian kernel for $\boldsymbol{h}$. In this work, we have $L = s$ when $\boldsymbol{h}$ is a Gaussian kernel.
The coarse scale variations are computed by applying a differentiation operator. Hence, $\boldsymbol{a}$ can be constructed by convolving a derivative kernel $\boldsymbol{b}$ with a low-pass kernel $\boldsymbol{h}$,
\begin{equation}
\boldsymbol{a} = \boldsymbol{h} * \boldsymbol{b}.
\label{eqn:derivativeSSATV1}
\end{equation}
We choose $\boldsymbol{b} = [1, -1]$, which results in the forward-difference, and hence $\mathcal{\tilde{D}}_y\boldsymbol{f}_{x,y,z} = \boldsymbol{f'}_{x, y, z} - \boldsymbol{f'}_{x, y-1, z}$.~Note that the $l_1$ norm of $\boldsymbol{b} = [1, -1]$ (Eqn.~(\ref{eqn:gradientOperator})) is 2 and the amplitude of $\mathcal{\tilde{D}}_y\boldsymbol{f}_{x,y,z}$ determines the weight $\boldsymbol{w}^{(n)}_{x,y,z}$ which influences the convergence speed and image resolution. Therefore, the combined kernel $\boldsymbol{a}$ is also scaled to have an $l_1$ norm of 2.

In analogy to Eqn.~(\ref{eq:TVgradient}), the partial derivative of $||\boldsymbol{f}||_{\text{wTV}}$ w.\,r.\,t.~voxel $\boldsymbol{f}_{x,y,z}$ then turns out to be,
\begin{equation}
\centering
\begin{split}
&\mathbf{g}_{x,y,z}= 
\sum_{i=-s}^{s-1}\left(\boldsymbol{w}^{(n)}_{x,y-i,z}\cdot \frac{\boldsymbol{a}_{s+i}\cdot \mathcal{\tilde{D}}_y\boldsymbol{f}_{x,y-i,z}}{||\mathcal{\tilde{D}}\boldsymbol{f}_{x,y-i,z}||}\right)\\
&+\boldsymbol{w}^{(n)}_{x,y,z}\cdot \frac{\mathcal{D}_x\boldsymbol{f}_{x,y,z} + \mathcal{D}_z\boldsymbol{f}_{x,y,z}}{||\mathcal{\tilde{D}}\boldsymbol{f}_{x,y,z}||}
-\boldsymbol{w}^{(n)}_{x+1,y,z}\cdot \frac{\mathcal{D}_x\boldsymbol{f}_{x+1,y,z}}{||\mathcal{\tilde{D}}\boldsymbol{f}_{x+1,y,z}||}\\
&-\boldsymbol{w}^{(n)}_{x,y,z+1}\cdot \frac{\mathcal{D}_z\boldsymbol{f}_{x,y,z+1}}{||\mathcal{\tilde{D}}\boldsymbol{f}_{x,y,z+1}||}.
\end{split}
\label{eqn:ssTVgradient}
\end{equation}

The general framework of ssaTV-1 is shown in Fig.~\ref{fig:ssaTValgorithm}. For each scale $s$, the ssaTV minimization box is the same as the wTV regularization box in Fig.~\ref{fig:wTValgorithm} but uses the modified derivative-like operator $\mathcal{\tilde{D}}_y\boldsymbol{f}$.
We choose the scales $s\in \left\lbrace 2^{l_{\max}-1}, 2^{l_{\max}-2}, \ldots, 2^{l}, \ldots, 1\right\rbrace$ with decreasing powers of two like classic Gaussian pyramids \cite{adelson1984pyramid} where $l_{\max}$ is the maximum level used. The standard deviation of the Gaussian kernel at each scale $s$ is chosen as $\sigma_s = \sqrt{s/2}$ \cite{adelson1984pyramid}.
Note that for $s=1$ a regular wTV regularization step is used.

\subsubsection{The second ssaTV algorithm (ssaTV-2)}
\begin{figure*}[t]
\centering
\begin{minipage}[b]{0.8\linewidth}
  \includegraphics[width=\textwidth]{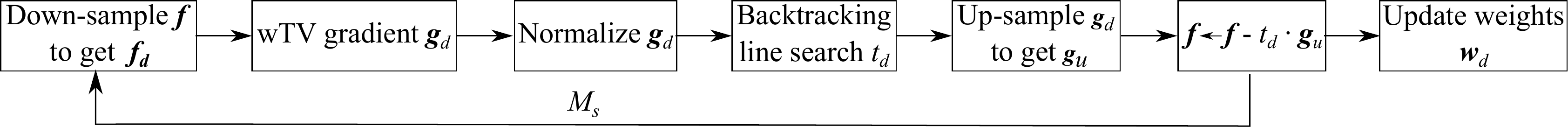}
\end{minipage}
 \caption{The ssaTV-2 minimization substep down-samples the image $\boldsymbol{f}$ to calculate the wTV gradient $\mathbf{g}_d$ and step size $t_d$, then it uses $t_d$ and the up-sampled $\mathbf{g}_u$ to update the original image $\boldsymbol{f}$.}
  \label{fig:ssTVMinimization}
\end{figure*}

In ssaTV-1, we increase the scale of the TV regularization with an anisotropic gradient operator while the scales of the streak artifacts remain the same. As an alternative, in ssaTV-2, the gradient operator is kept unchanged while the size of the image, and thus also of the streak artifacts, is reduced by down-sampling.

Like ssaTV-1, the down-sampling operation is also applied anisotropically in $Y$ direction only. It is defined as the sub-sampling of the low-pass-filtered image,
\begin{equation}
\left(\boldsymbol{f}_d\right)_{x, y', z} = \boldsymbol{f'}_{x, s \cdot y' , z},
\label{eq:downSamplingOperation}
\end{equation}
where $\boldsymbol{f'}$ is defined in Eqn.~(\ref{eqn:lowPassFiltering}), $\boldsymbol{f}_d$ is the down-sampled image, and $y'$ is the $Y$-index of $\boldsymbol{f}_d$.

On the down-sampled image, the partial derivative of $||\boldsymbol{f}_d||_{\text{wTV}}$ w.\,r.\,t.~each voxel $\left(\boldsymbol{f}_d\right)_{x, y', z}$ can be conveniently calculated like Eqn.~(\ref{eq:TVgradient}), denoted by $\mathbf{g}_d$,
\begin{equation}
\left(\mathbf{g}_d\right)_{x, y', z}=\frac{\partial{||\boldsymbol{f}_d||_{\text{wTV}}}}{\partial {\left(\boldsymbol{f}_d\right)_{x, y', z}}}.
\end{equation}

In order to reduce low frequency streaks at the original image, the partial derivative of $||\boldsymbol{f}_d||_{\text{wTV}}$ w.\,r.\,t.~each voxel of the original image $\boldsymbol{f}$ is needed, denoted by $\mathbf{g}_u$,
\begin{equation}
\left(\mathbf{g}_u\right)_{x, y, z}=\frac{\partial{||\boldsymbol{f}_d||_{\text{wTV}}}}{\partial {\boldsymbol{f}_{x, y, z}}}.
\end{equation}

Applying the multi-variant chain rule, $\mathbf{g}_u$ and $\mathbf{g}_d$ have the following relation,
\begin{equation}
\begin{split}
&(\mathbf{g}_u)_{x,y,z}
=\frac{\partial{||\boldsymbol{f}_d||_{\text{wTV}}}}{\partial {\boldsymbol{f}_{x,y,z}}}
\\
&=\sum_{k}\frac{\partial{||\boldsymbol{f}_d||_{\text{wTV}}}}{\partial {(\boldsymbol{f}_d)_{x, y'+k,z}}} \cdot 
\frac{\partial {(\boldsymbol{f}_d)_{x, y'+k,z}}}{\partial {\boldsymbol{f}_{x,y,z}}}|_{y=s\cdot y'+j}\\
&= \sum_{k}(\mathbf{g}_d)_{x,y'+k,z} \cdot \boldsymbol{h}_{L-j+s \cdot k},
\end{split}
\end{equation}
where $j \in \left\lbrace -L, -L+1,\ldots, L\right\rbrace$, $k\in \left\lbrace 0,\pm 1, \pm2, \ldots\right\rbrace$. Here $\boldsymbol{h}_{i} = 0$ when $i < 0$ or $i > 2\cdot L$. Therefore, $\mathbf{g}_u$ can be obtained from $\mathbf{g}_d$ by the following up-sampling operation,
\begin{equation}
\begin{split}
&(\mathbf{g}'_u)_{x,s\cdot y'+j,z} = \left\{
\begin{array}{l}
(\mathbf{g}_d)_{x,y',z}, \quad j=0,\\
0, \quad  j\in \left\lbrace\pm 1, \pm 2,\ldots, \pm L\right\rbrace,
\end{array}
\right.\\
&(\mathbf{g}_u)_{x, y, z} = \sum_{j = -L}^{L} {\boldsymbol{h'}_{L-j}\cdot (\mathbf{g}'_u)_{x, y + j, z}},
\end{split}
\end{equation}
where $\mathbf{g}'_u$ is up-sampled from $\mathbf{g}_d$ by inserting zeros between samples, $\boldsymbol{h'}$ is the reversed $\boldsymbol{h}$, i.\,e., $\boldsymbol{h'}_j=\boldsymbol{h}_{2L-j}, j \in \left\lbrace 0, 1, \dots, 2L \right\rbrace$, and $\mathbf{g}_u$ is the convolution of $\mathbf{g}'_u$ and $\boldsymbol{h'}$ \cite{adelson1984pyramid}.

The ssaTV-2 algorithm also follows the general framework of Fig.~\ref{fig:ssaTValgorithm}, with regularization at each scale replaced by the ssaTV-2 minimization substep shown in Fig.~\ref{fig:ssTVMinimization}. Each substep first down-samples the image $\boldsymbol{f}$ with the scaling factor $s$ to obtain $\boldsymbol{f}_d$, on which the partial derivative $\mathbf{g}_d$ is calculated and a suitable step size $t_d$ is found by backtracking line search such that the wTV value of $\boldsymbol{f}_d-t_d\cdot\mathbf{g}_d$ decreases.
Subsequently, $\mathbf{g}_d$ is up-sampled with the same scaling factor $s$ to obtain $\mathbf{g}_u$. Finally, the original image $\boldsymbol{f}$ is updated as $\boldsymbol{f} \gets \boldsymbol{f}-t_d\cdot\mathbf{g}_u$. The above process is repeated $M_s$ times, then the corresponding weights $\boldsymbol{w}_d$ are updated.

\subsection{Experimental setup}
\subsubsection{Numerical phantom}
\begin{figure}[t]
\centering
\begin{minipage}[t]{0.3\linewidth}
\centering
\subfigure[FORBILD phantom]{
\includegraphics[width = 1\linewidth]{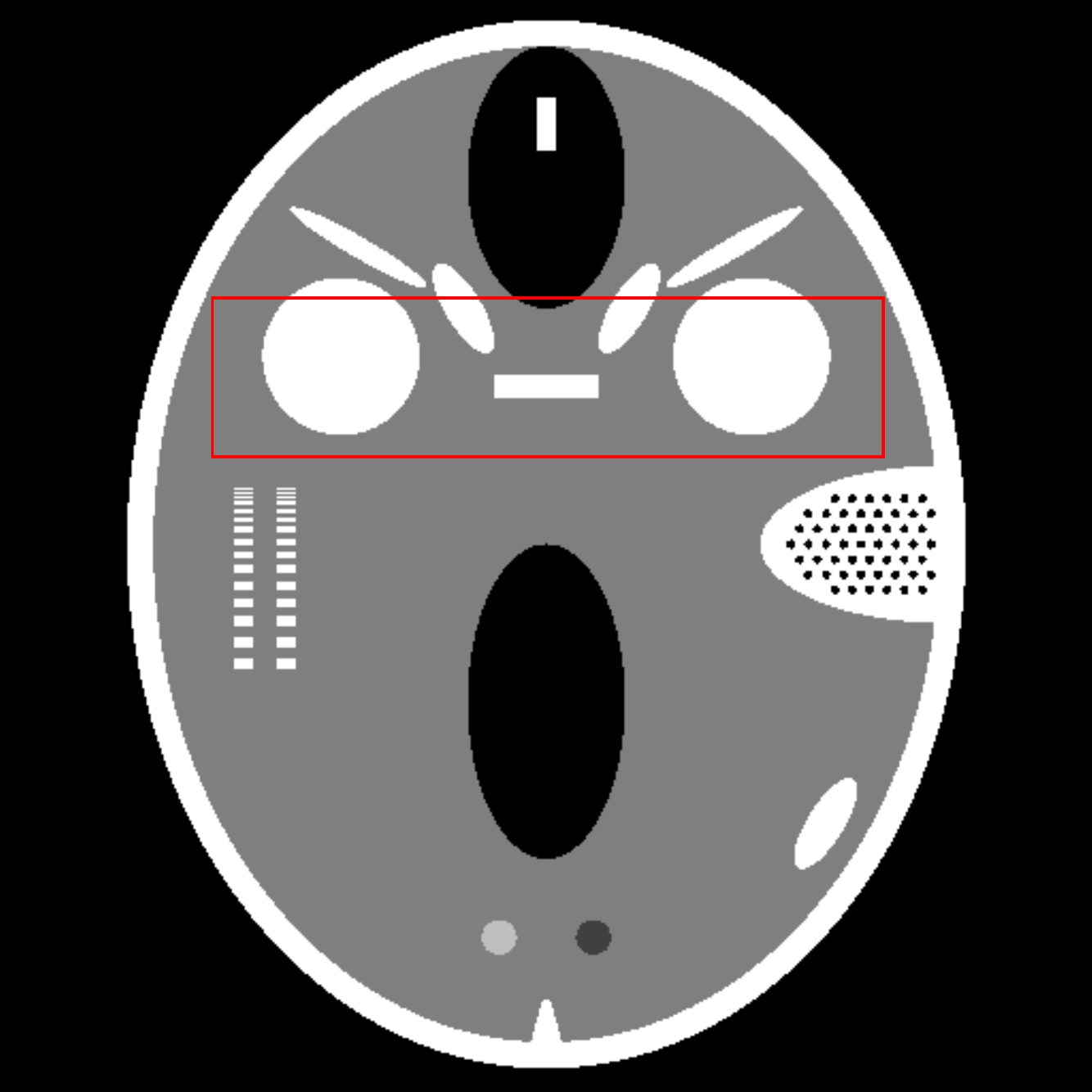}
\label{subfig:ForbildPhantom}
}
\end{minipage}
\begin{minipage}[t]{0.3\linewidth}
\centering
\subfigure[SART]{
\includegraphics[width = 1\linewidth]{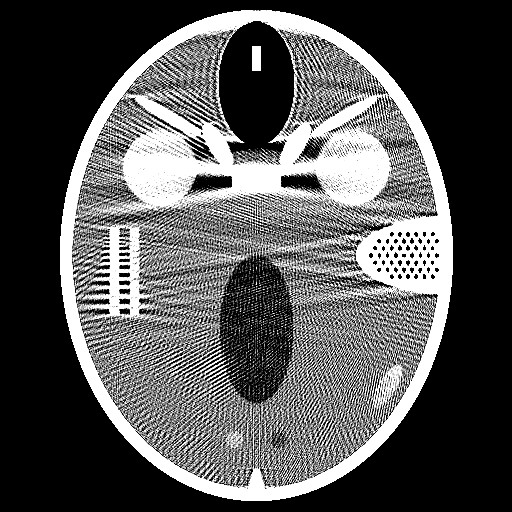}
\label{subfig:SARTPhantom}
}
\end{minipage}
\caption{The modified FORBILD phantom and its SART reconstruction from the $160^\circ$ limited angle sinogram. The red box is the region of interest. Window: [0, 100]\,HU.}
\end{figure}


In order to validate the advantage of both ssaTV algorithms in reducing large streaks, experiments on a modified pixelized 2-D FORBILD phantom (Fig.~\ref{subfig:ForbildPhantom}) \cite{yu2012simulation} are performed. The image size is $512 \times 512$ pixels with an isotropic pixel size of $0.5$\,mm. The original attenuation coefficients are found in \cite{headPhantom}. As we do not expect a useful low-contrast performance in the limited-angle scenario and thus prefer to focus on objects with moderate or high contrast, structures with a contrast of 2.5\,HU, 5\,HU and 10\,HU are modified to 25\,HU, 50\,HU, and 100\,HU. To test the spatial resolution of reconstructed images in $Y$ direction, the original left ear of the phantom is replaced by two sequences of bars. Since we apply the scaling along $Y$ direction, the risk of losing spatial resolution is higher here. Therefore, we stack bars along this direction. The bars are categorized as high contrast and medium contrast bars with attenuation coefficients of $800\,$HU and $250\,$HU, respectively. Each bar sequence contains 5 triples with increasing width from $0.5\,$mm to $2.5\,$mm at an increment of $0.5\,$mm. The space between bars is equal to the width of the bar. The length of all bars is $4.5\,$mm.

For 2-D numerical experiments, a fan-beam scan is simulated with the trajectory shown in Fig.~\ref{fig:trajectory}. The scan angle range from $10^\circ$ to $170^\circ$ is chosen such that most streaks are expected in the horizontal direction. The size of the equal-spaced detector is $768$ pixels and the detector element size is $0.5\,$mm. The source to detector distance is $d=1088\,$mm, the source to isocenter distance is $d' = d/2 = 544\,$mm, the fan angle is $\gamma_{\max}=20^\circ$ and the trajectory angular increment is $1^\circ$. Two experiments are performed, with and without Poisson noise. The Poisson noise is simulated considering a total number of $5\cdot 10^6$ incident photons at each detector pixel without object attenuation. The simulated X-rays are mono-energetic at 65\,KeV. A linear attenuation coefficient of 0.02/mm is chosen as 0\,HU. No scattering is considered.

\subsubsection{Clinical data}
The proposed algorithms are also evaluated on a 3-D clinical head dataset with a typical noise level acquired from an Artis zee angiographic C-arm system (Siemens Healthcare GmbH, Forchheim, Germany). The dose area product (DAP) of the complete scan is $532\,$\textmu Gy$\cdot$m$^2$. The detector size is 1240$\times$960 pixels with an isotropic detector pixel size of $0.308\,$mm. The complete dataset contains $496$ projections obtained in a $200^\circ$ short scan.
We use wTV to reconstruct the complete data as an image quality reference. The reconstructed image is $512\times 512\times 256$ voxels large, with a voxel size of $0.4\,$mm, $0.4\,$mm, and $0.8\,$mm in $X$, $Y$, and $Z$, respectively.

For the limited angle setting, we simulate three acquisitions with angular ranges covering $160^\circ$, $140^\circ$, and $120^\circ$. The angular ranges are $10^\circ-170^\circ$, $20^\circ-160^\circ$, and $30^\circ-150^\circ$ and obtained by keeping only the projection images 50 through 446, 75 through 421, and 100 through 396, respectively.   

\subsubsection{Reconstruction parameters}
From the limited angle projections of the numerical phantom and the clinical data, images are reconstructed with SART, wTV, and both versions of ssaTV. SART is used to show the image quality without TV regularization (Fig.~\ref{subfig:SARTPhantom}). We choose the relaxation parameter $\lambda=0.8$ in Eqn.~(\ref{eqn:SART}). For wTV, we choose $M=10$ gradient descent steps, where the backtracking line search uses a gradient shrink parameter $\alpha = 0.3$ and a step size update parameter $\beta = 0.6$ \cite{boyd2004convex}. For both ssaTV algorithms, the same number of total gradient descent steps are applied for a fair comparison, i.\,e.~$\sum_{l=0}^{l_{\max}-1} M_{2^l}=M=10$. With this constraint, different combinations of $M_s$ are possible, which influence the convergence behavior. For example, when $l_{\max}=2$, we have $M_1+M_2=M=10$. If we choose a larger number for $M_2$, large streaks are reduced faster. However, when $M_1$ is too low, small streaks and high frequency noise may not be reduced effectively. Therefore, the chosen combination of $M_s$ is a trade-off. In this paper, empirically the following combinations are investigated: $[M_1, M_2, M_4, M_8, M_{16}] = [5, 5, 0, 0, 0], [3, 3, 4, 0, 0], [3, 3, 2, 2, 0], [2, 2, 2, 2, 2]$ for $l_{\max} = 2, 3, 4, 5$, respectively.

For both ssaTV algorithms, we use the normalized binomial coefficients as approximations of the Gaussian kernel $\boldsymbol{h}$ \cite{lindeberg1990scale}, i.\,e., $\boldsymbol{h}_j = {2\cdot s \choose j}/2^{2\cdot s}$ for scale $s$, which has a standard deviation $\sigma_s = \sqrt{s/2}$. 

The parameter $\epsilon$ in Eqn.~(\ref{eq:WeightsUpdate}) can be chosen in the range of 1\,HU and 50\,HU. A smaller value of $\epsilon$ leads to higher image resolution but a slower convergence speed \cite{Candes}. In our experiments, we choose $\epsilon = 5\,$HU for the FORBILD phantom. Each algorithm is run for 500 iterations. For the 3-D clinical experiments, the images are reconstructed with $\epsilon = 20$\,HU. In the clinical case, we only run the algorithms for 100 iterations as we observe no significant image quality improvement beyond that point. In both experiments, optimization is initialized with zero images.

\subsubsection{Quality metrics}
\label{subsubsection:qualityMetrics}
As a quality metric for the numerical experiment, we compute the root-mean-square error (RMSE) in a region of interest (ROI). The ROI is chosen to cover the area between the eyes where we expect that most large streaks occur (Fig.~\ref{subfig:ForbildPhantom}).

 The whole experimental setup is implemented in CONRAD, a software framework for medical imaging processing \cite{maier2013conrad}.

\section{Results}
\subsection{Numerical results}
\begin{figure}
\raggedright
\begin{minipage}[b]{0.95\linewidth}
\centering
\subfigure[ROI RMSE plots for ssaTV-1]{
  \centerline{\includegraphics[width=\textwidth, height = 130pt]{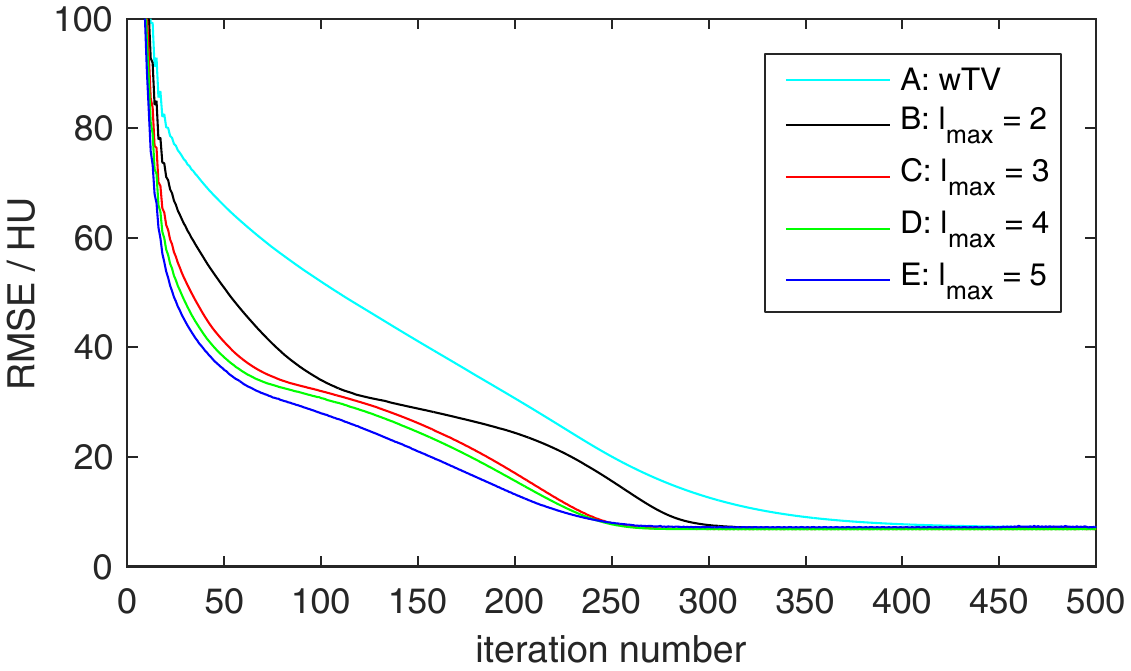}}
  }
\end{minipage}
\hfill
\begin{minipage}[b]{0.95\linewidth}
\centering
\subfigure[ROI RMSE plots for ssaTV-2]{
  \centerline{\includegraphics[width=\textwidth,height=130pt]{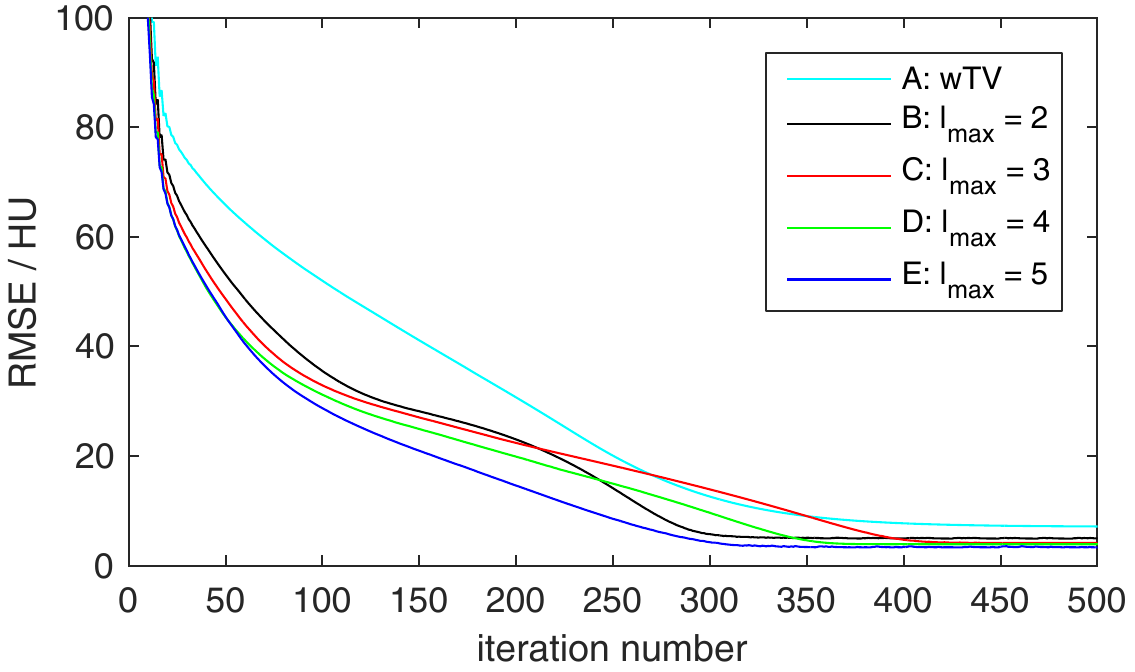}}
  }
\end{minipage}
\caption{ROI RMSE plots for ssaTV-1 and ssaTV-2 at different scaling levels on the FORBILD phantom without noise.}
\label{Fig:convergenceIndicators}
\end{figure}
The RMSE at the ROI for ssaTV-1 and ssaTV-2 at different scaling levels is plotted in Fig.~\ref{Fig:convergenceIndicators}. The ROI RMSE of wTV converges after about 400 iterations to the value of 7.0\,HU. The ROI RMSE of ssaTV-1 with different scaling levels all converge to the same value but have faster convergence speed than that of wTV. Particularly, a high scaling level accelerates the speed of streak reduction (Fig.~\ref{Fig:convergenceIndicators}(a)), which demonstrates the advantage of our scale-space approach. The ROI RMSE of ssaTV-2 converge to 5.0\,HU, 4.3\,HU, 3.9\,HU, and 3.4\,HU for scaling levels 2 through 5 respectively after about 400 iterations. Generally, they converge faster than that of wTV as well, with a single exception in the case of $l_{\max} = 3$ near the 300th iteration.

\begin{figure}[t]
\begin{minipage}[b]{0.3\linewidth}
\subfigure[wTV]{
\includegraphics[width = \textwidth]{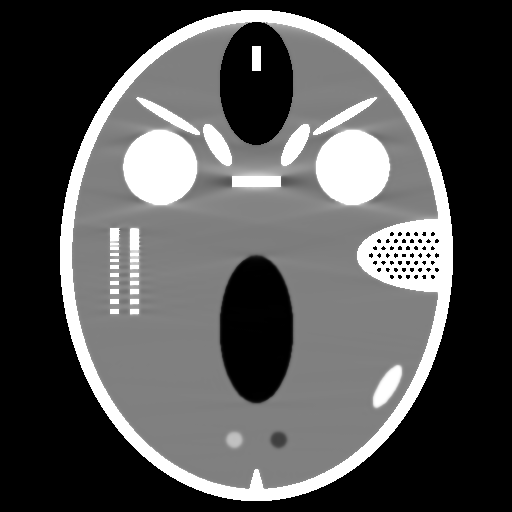}
}
\end{minipage}
\begin{minipage}[b]{0.3\linewidth}
\subfigure[ssaTV-1]{
\includegraphics[width = \textwidth]{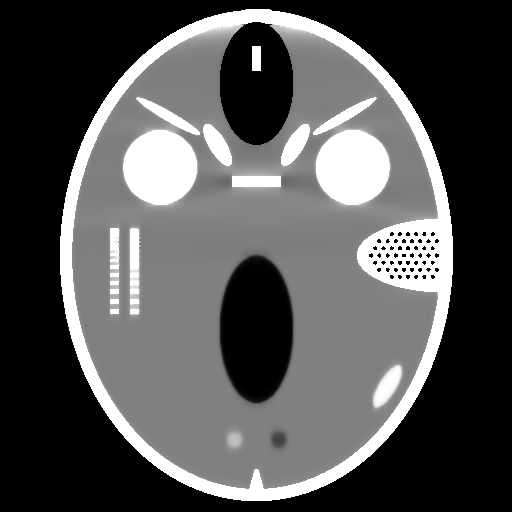}
}
\end{minipage}
\begin{minipage}[b]{0.3\linewidth}
\subfigure[ssaTV-2]{
\includegraphics[width = \textwidth]{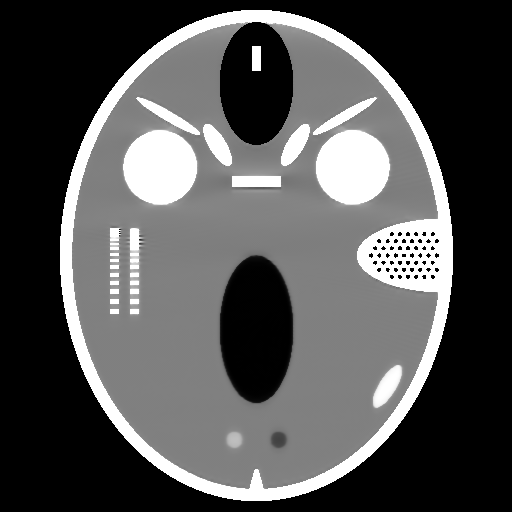}
}
\end{minipage}

\begin{minipage}[b]{0.3\linewidth}
\subfigure[wTV, difference]{
\includegraphics[width = \textwidth]{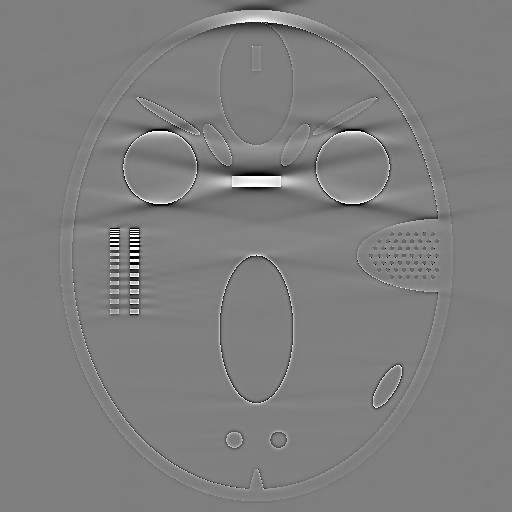}
}
\end{minipage}
\begin{minipage}[b]{0.3\linewidth}
\subfigure[ssaTV-1, difference]{
\includegraphics[width = \textwidth]{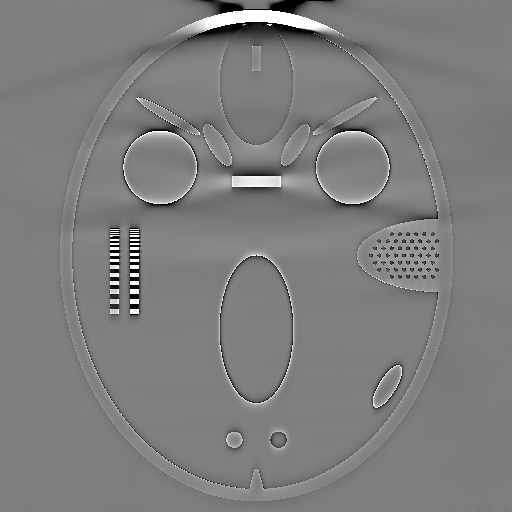}
}
\end{minipage}
\begin{minipage}[b]{0.3\linewidth}
\subfigure[ssaTV-2, difference]{
\includegraphics[width = \textwidth]{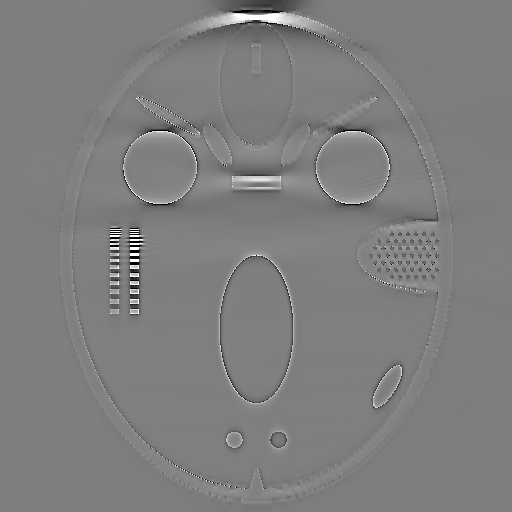}
}
\end{minipage}

\begin{minipage}[b]{0.3\linewidth}
\subfigure[wTV, bars]{
\includegraphics[width = \textwidth]{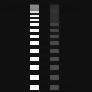}
}
\end{minipage}
\begin{minipage}[b]{0.3\linewidth}
\subfigure[ssaTV-1, bars]{
\includegraphics[width = \textwidth]{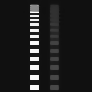}
}
\end{minipage}
\begin{minipage}[b]{0.3\linewidth}
\subfigure[ssaTV-2, bars]{
\includegraphics[width = \textwidth]{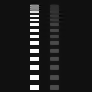}
}
\end{minipage}
\caption{Comparison of wTV, ssaTV-1 ($l_{\max}=3$), and ssaTV-2 ($l_{\max}=3$) using the modified FORBILD phantom without noise, 500 iterations. Window: [0 100]\,HU for the top row, [0 800]\,HU for the bottom row, and a window width of 50\,HU for the difference images at the middle row.}
\label{Fig:ForbildResults}
\end{figure}

The final reconstruction results (exemplarily for $l_{\max} = 3$) and their differences from the ground truth are shown in Fig.~\ref{Fig:ForbildResults}. The reconstructed images (Figs.~\ref{Fig:ForbildResults}(a)-(c)) demonstrate that both ssaTV algorithms reduce streaks better than wTV. As a side remark, the difference images (Figs.~\ref{Fig:ForbildResults}(d)-(f)) reveal that all three algorithms fail to recover the top boundary where most data is missing \cite{huang2016image}.

Regarding image resolution, Figs.~\ref{Fig:ForbildResults}(g)-(i) show that both ssaTV algorithms reconstruct high contrast bars as well as wTV does. For all algorithms, the finest bars are blurred due to missing data. For the medium contrast bars, ssaTV-2 separates them better than wTV and ssaTV-1. 

\begin{figure}[t]
\begin{minipage}[b]{0.3\linewidth}
\subfigure[wTV]{
\includegraphics[width = \textwidth]{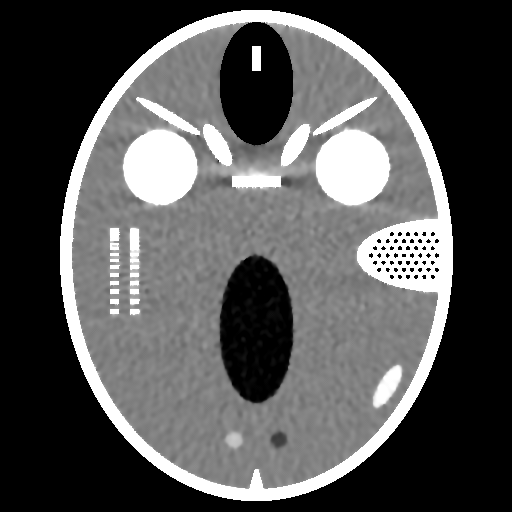}
}
\end{minipage}
\begin{minipage}[b]{0.3\linewidth}
\subfigure[ssaTV1]{
\includegraphics[width = \textwidth]{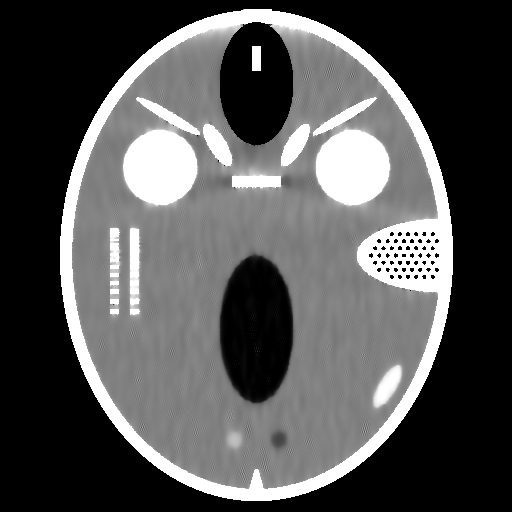}
}
\end{minipage}
\begin{minipage}[b]{0.3\linewidth}
\subfigure[ssaTV2]{
\includegraphics[width = \textwidth]{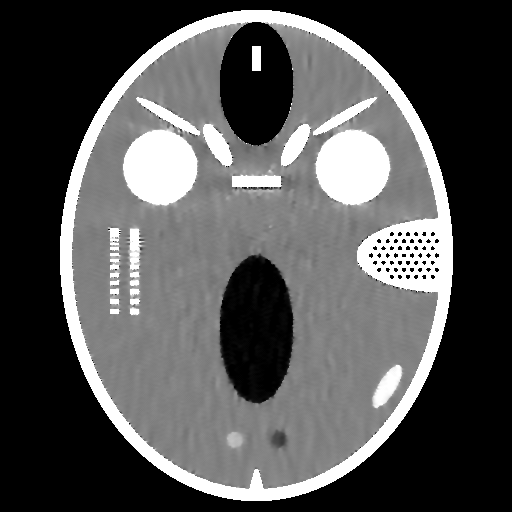}
}
\end{minipage}
\caption{Comparison of wTV, ssaTV-1 ($l_{\max}=3$), and ssaTV-2 ($l_{\max}=3$) using the modified FORBILD phantom with Poisson noise, 500 iterations. Window: [0 100]\,HU.}
\label{Fig:ForbildNoisy}
\end{figure}

The reconstruction results for the FORBILD phantom with Poisson noise are displayed in Fig.~\ref{Fig:ForbildNoisy}. Both ssaTV algorithms can handle noisy data and show its superiority in streak artifact reduction. It has to be noted that the anisotropic scaling approach smoothes noise in an anisotropic manner as well.
\subsection{Clincial Results}
\begin{figure}[t]
\centering
\begin{minipage}[b]{0.3\linewidth}
\centering
\subfigure[wTV, 65th slice]{
  \centerline{\includegraphics[width=\textwidth]{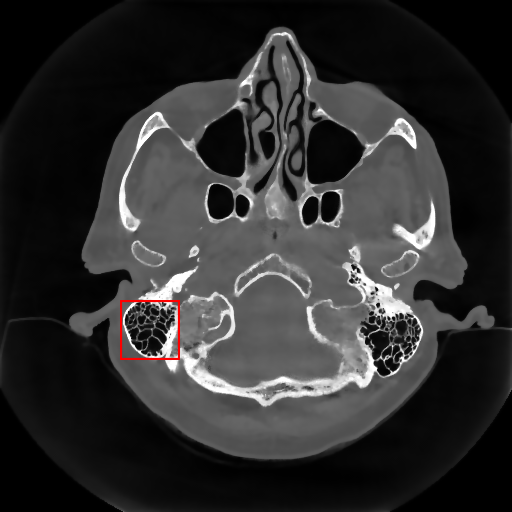}}
  \label{subfig:clinicReferenceBoneWithROI}
  }
\end{minipage}
\begin{minipage}[b]{0.3\linewidth}
\centering
\subfigure[wTV, 140th slice]{
  \centerline{\includegraphics[width=\textwidth]{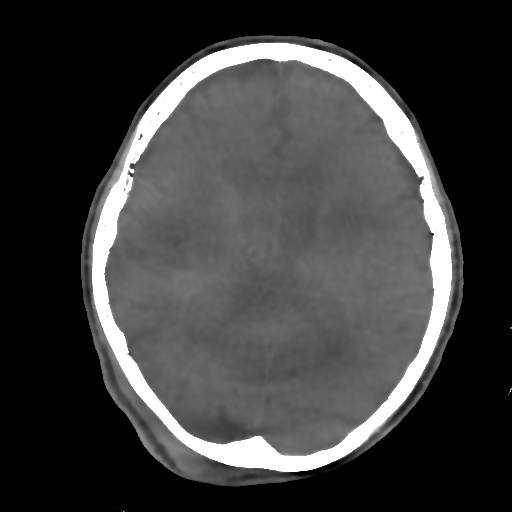}}
  }
\end{minipage}
\caption{Reference images reconstructed from the complete clinical dataset with wTV. The red box is the ROI for fine bone structures. Window: [-1000 1730] HU and [-220 365] HU for left and right images, respectively.}
\label{Fig:wTVreference}
\end{figure}
\begin{figure}
\begin{minipage}[b]{0.24\linewidth}
\centering
\subfigure[SART]{
  \centerline{\includegraphics[width=\textwidth]{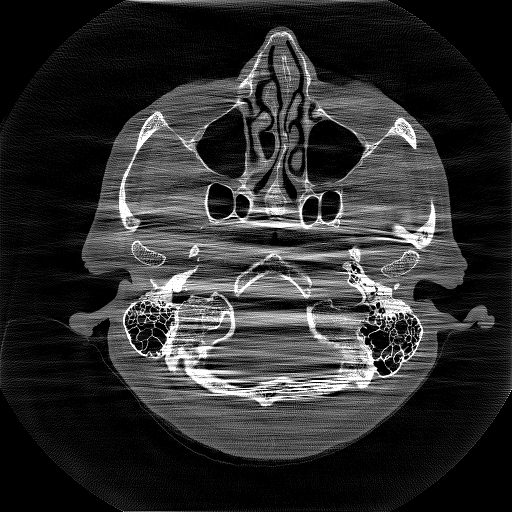}}
  \label{subfig:SARTclinicBone}
  }
\end{minipage}
\begin{minipage}[b]{0.24\linewidth}
\centering
\subfigure[SART]{
  \centerline{\includegraphics[width=\textwidth]{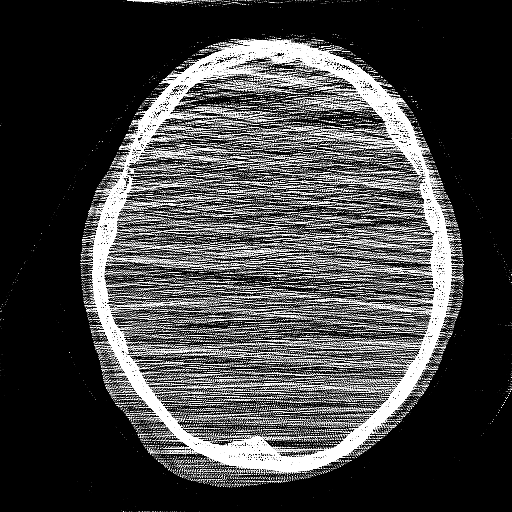}}
 \label{subfig:SARTclinicBrain}
}
\end{minipage}
\begin{minipage}[b]{0.24\linewidth}
\centering
\subfigure[difference]{
  \centerline{\includegraphics[width=\textwidth]{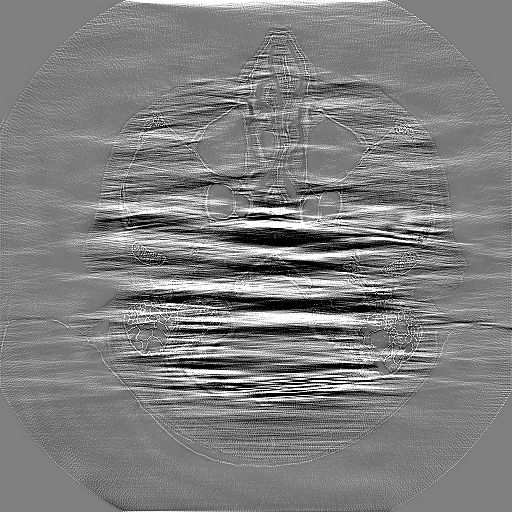}}
  }
\end{minipage}
\begin{minipage}[b]{0.24\linewidth}
\centering
\subfigure[difference]{
  \centerline{\includegraphics[width=\textwidth]{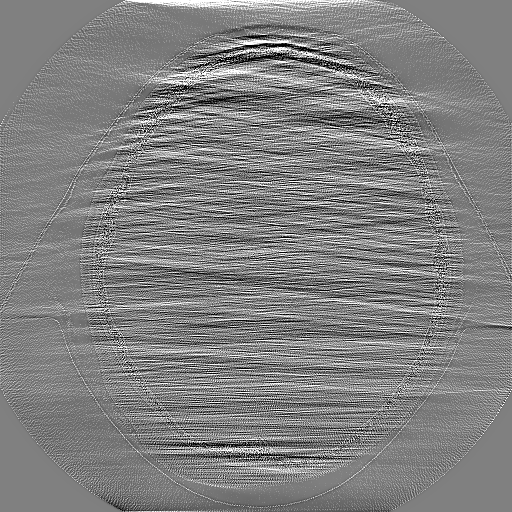}}
}
\end{minipage}
 
\begin{minipage}[b]{0.24\linewidth}
\centering
\subfigure[wTV]{
  \centerline{\includegraphics[width=\textwidth]{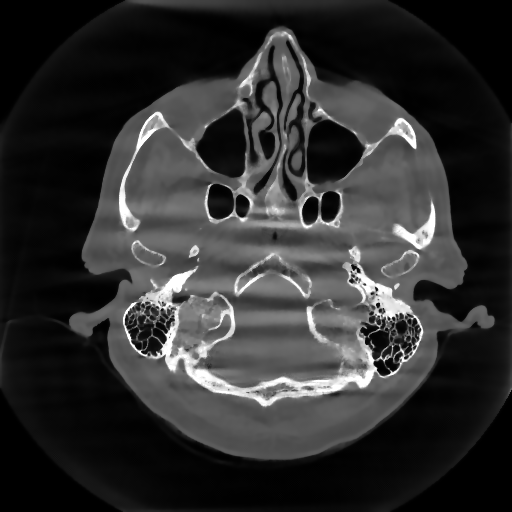}}
  \label{subfig:wTVclinicBone}
}
\end{minipage}
\begin{minipage}[b]{0.24\linewidth}
\centering
\subfigure[wTV]{
  \centerline{\includegraphics[width=\textwidth]{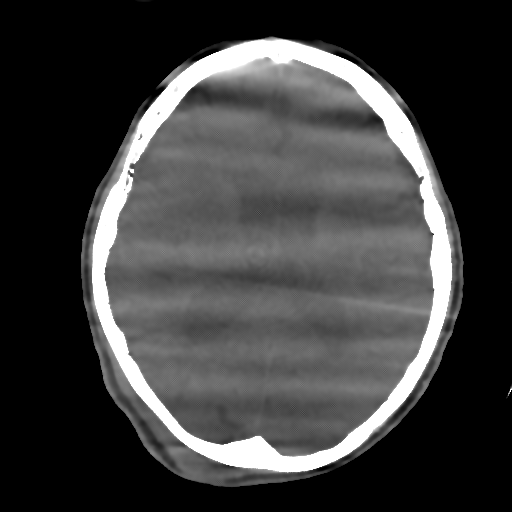}}
  \label{subfig:wTVclinicBrain}
}
\end{minipage}
\begin{minipage}[b]{0.24\linewidth}
\centering
\subfigure[difference]{
  \centerline{\includegraphics[width=\textwidth]{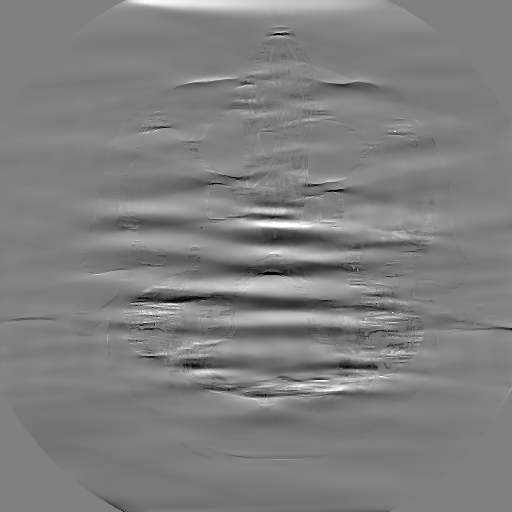}}
}
\end{minipage}
\begin{minipage}[b]{0.24\linewidth}
\centering
\subfigure[difference]{
  \centerline{\includegraphics[width=\textwidth]{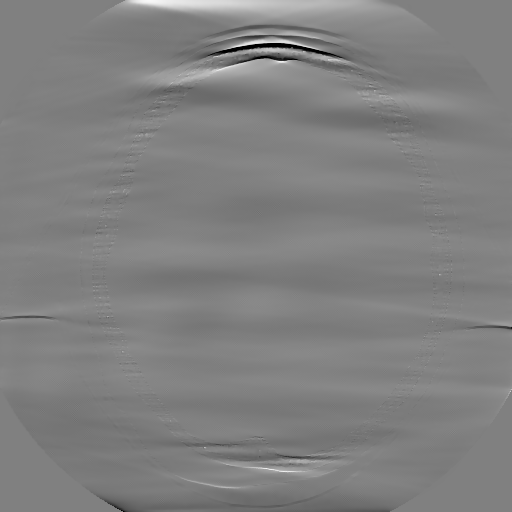}}
}
\end{minipage}

\begin{minipage}[b]{0.24\linewidth}
\centering
\subfigure[ssaTV-1]{
  \centerline{\includegraphics[width=\textwidth]{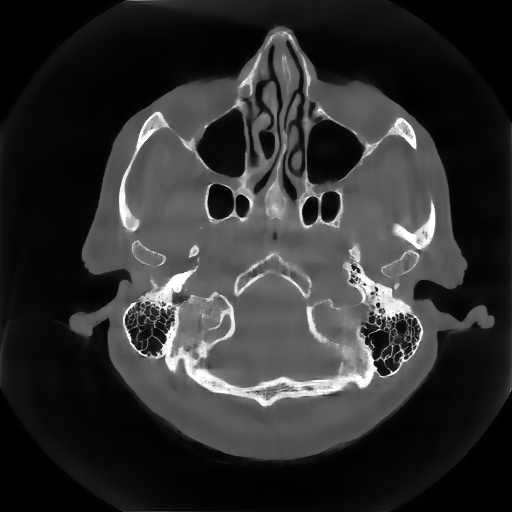}}
  }
\end{minipage}
\begin{minipage}[b]{0.24\linewidth}
\centering
\subfigure[ssaTV-1]{
  \centerline{\includegraphics[width=\textwidth]{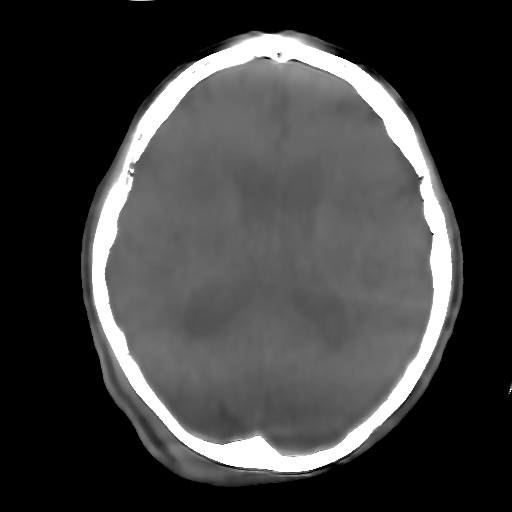}}
\label{subfig:ssaTV1Bone} 
  }
\end{minipage}
\begin{minipage}[b]{0.24\linewidth}
\centering
\subfigure[difference]{
  \centerline{\includegraphics[width=\textwidth]{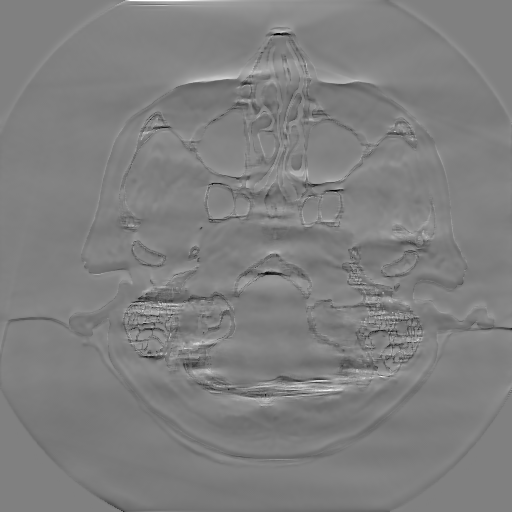}}
  }
\end{minipage}
\begin{minipage}[b]{0.24\linewidth}
\centering
\subfigure[difference]{
  \centerline{\includegraphics[width=\textwidth]{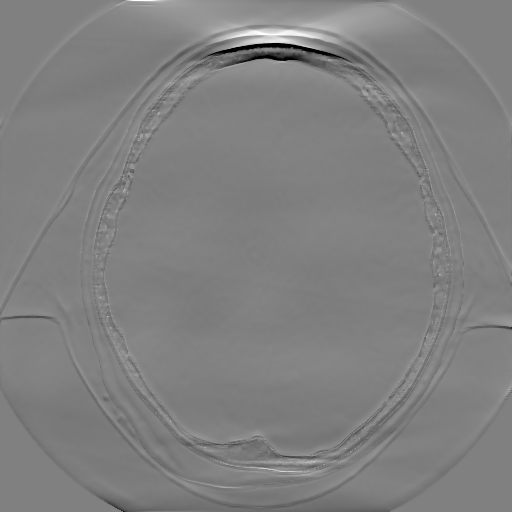}}
  }
\end{minipage}

\begin{minipage}[b]{0.24\linewidth}
\centering
\subfigure[ssaTV-2]{
  \centerline{\includegraphics[width=\textwidth]{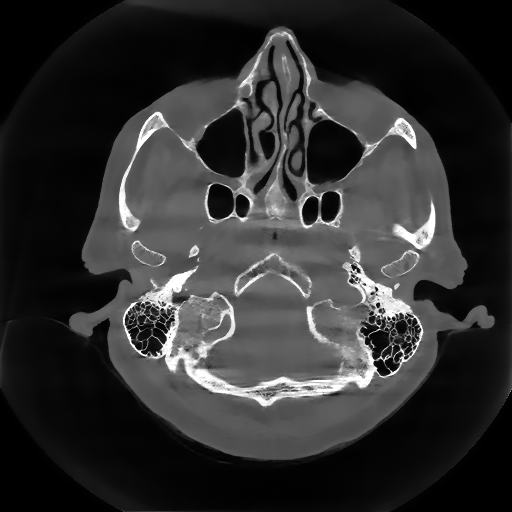}}
  }
\end{minipage}
\begin{minipage}[b]{0.24\linewidth}
\centering
\subfigure[ssaTV-2]{
  \centerline{\includegraphics[width=\textwidth]{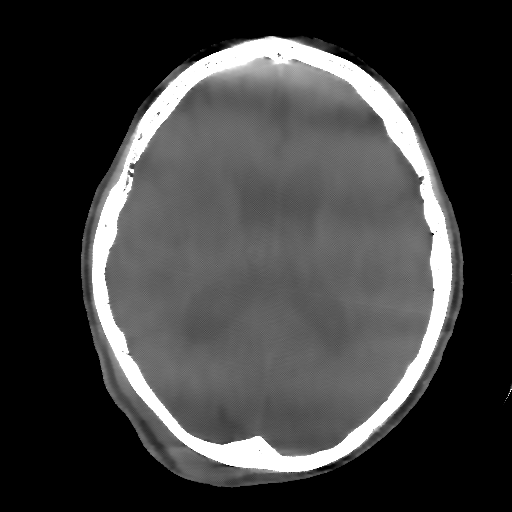}}  
  \label{subfig:ssaTV2Bone}
  }  
\end{minipage}
\begin{minipage}[b]{0.24\linewidth}
\centering
\subfigure[difference]{
  \centerline{\includegraphics[width=\textwidth]{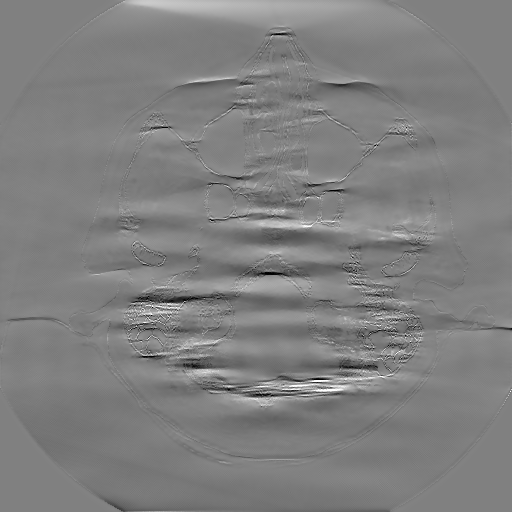}}
  }
\end{minipage}
\begin{minipage}[b]{0.24\linewidth}
\centering
\subfigure[difference]{
  \centerline{\includegraphics[width=\textwidth]{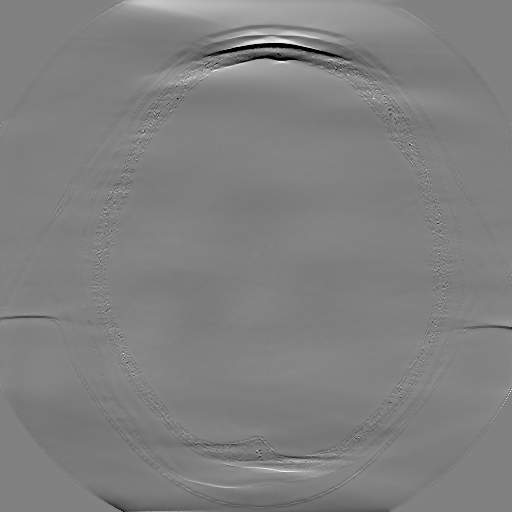}}
  }
\end{minipage}
\caption{Reconstruction results of SART, wTV, ssaTV-1 ($l_{\max} = 3$), and ssaTV-2 ($l_{\max} = 3$) from $160^\circ$ limited angle data, 100 iterations. Window: [-1000 1730] HU and [-220 365] HU for the first and second columns, respectively. The difference images at the third and fourth columns are shown with a window width of 780\,HU.}
\label{Fig:ComparisonOfAlgorithmsInClinical}
\end{figure}
\begin{figure}
\begin{minipage}[b]{0.24\linewidth}
\centering
\subfigure[reference image]{
  \centerline{\includegraphics[width=\textwidth]{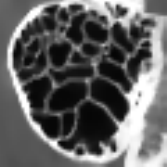}}
  }
\end{minipage}
\begin{minipage}[b]{0.24\linewidth}
\centering
\subfigure[wTV ]{
  \centerline{\includegraphics[width=\textwidth]{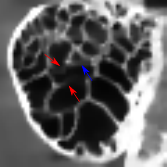}}
  \label{subfig:wTVROIclinic}
  }
\end{minipage}
\begin{minipage}[b]{0.24\linewidth}
\centering
\subfigure[ssaTV-1]{
  \centerline{\includegraphics[width=\textwidth]{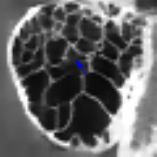}}
  }
\end{minipage}
\begin{minipage}[b]{0.24\linewidth}
\centering
\subfigure[ssaTV-2]{
  \centerline{\includegraphics[width=\textwidth]{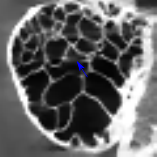}}
  }
\end{minipage}
\caption{Reconstructed fine bone structures for wTV, ssaTV-1 ($l_{\max}=3$), and ssaTV-2 ($l_{\max}=3$), window [-1000 1730]\,HU. The horizontal bone structures indicated by the arrows are missing or blurred.}
\label{fig:ROIclinic}
\end{figure}

Reference images of the complete clinical dataset reconstructed by wTV are shown in Fig.~\ref{Fig:wTVreference}.
The results of SART, wTV, ssaTV-1 ($l_{\max}$ = 3), and ssaTV-2 ($l_{\max} = 3$) reconstructed from $160^\circ$ limited angle data are shown in Fig.~\ref{Fig:ComparisonOfAlgorithmsInClinical}. Figs.~\ref{Fig:ComparisonOfAlgorithmsInClinical}(e) and (f) demonstrate that wTV removes small streaks and high frequency noise well. However, large streaks along the horizontal direction still exist and some anatomical structures are obscured by them. 
Compared to wTV, large streaks are reduced in both ssaTV results (Figs.~\ref{Fig:ComparisonOfAlgorithmsInClinical}(i)-(p)). Fig.~\ref{fig:ROIclinic} shows a zoom-in (red box in Fig.~\ref{Fig:wTVreference}) of the fine bone structures. Fig.~\ref{subfig:wTVROIclinic} displays that wTV fails to reconstruct some horizontal structures indicated by the red arrows that both ssaTV algorithms are able to recover better. 

\begin{figure}
\centering
\begin{minipage}[b]{0.3\linewidth}
\centering
\subfigure[wTV]{
  \centerline{\includegraphics[width=\textwidth]{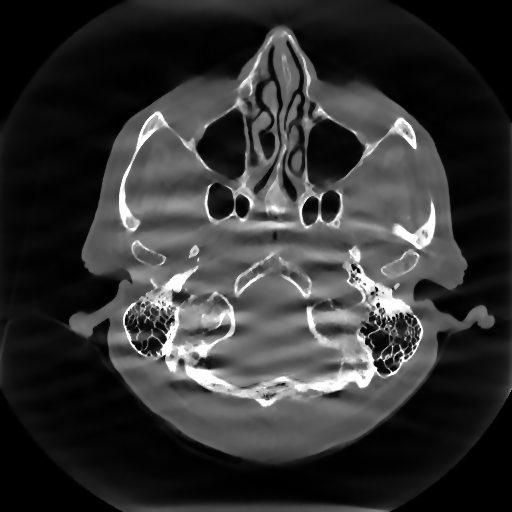}}
  }
\end{minipage}
\begin{minipage}[b]{0.3\linewidth}
\centering
\subfigure[ssaTV-1]{
  \centerline{\includegraphics[width=\textwidth]{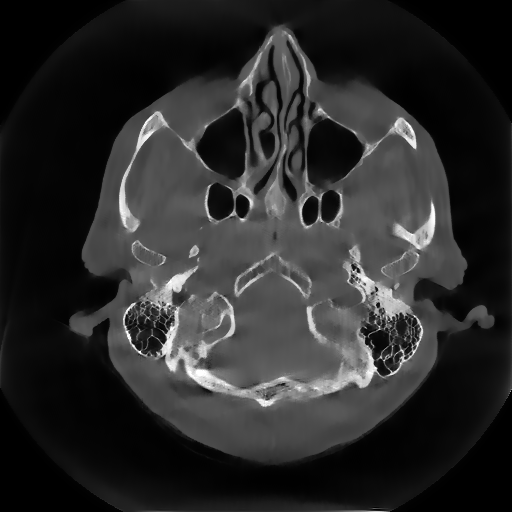}}
  }
\end{minipage}
\begin{minipage}[b]{0.3\linewidth}
\centering
\subfigure[ssaTV-2]{
  \centerline{\includegraphics[width=\textwidth]{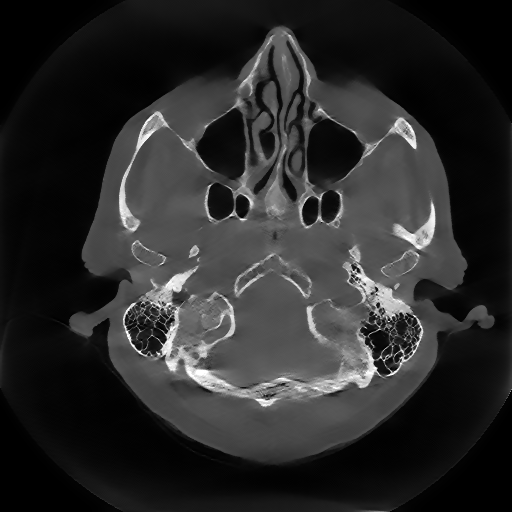}}
  }
\end{minipage}

\begin{minipage}[b]{0.3\linewidth}
\centering
\subfigure[wTV]{
  \centerline{\includegraphics[width=\textwidth]{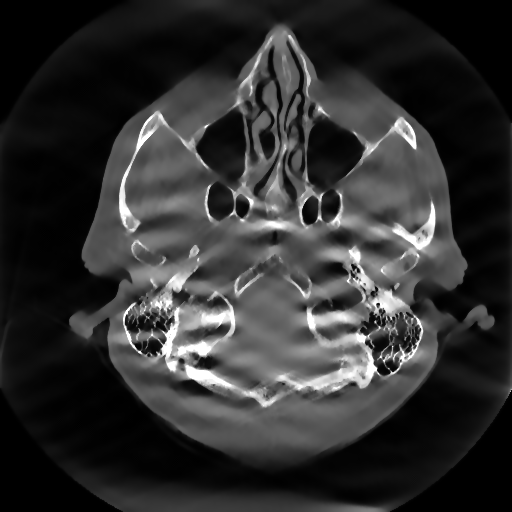}}
  }
  \end{minipage}
\begin{minipage}[b]{0.3\linewidth}
\centering
\subfigure[ssaTV-1]{
  \centerline{\includegraphics[width=\textwidth]{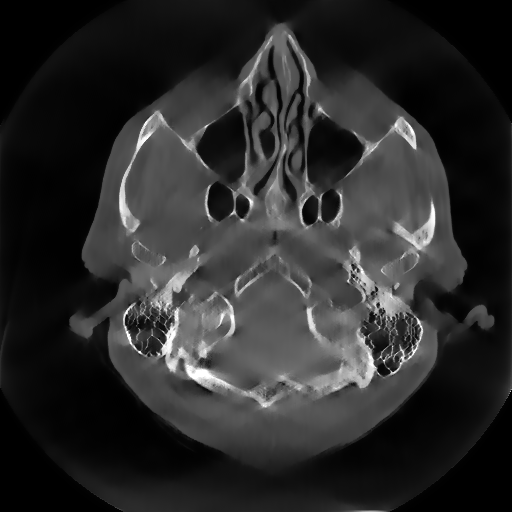}}
  }
\end{minipage}
\begin{minipage}[b]{0.3\linewidth}
\centering
\subfigure[ssaTV-2]{
  \centerline{\includegraphics[width=\textwidth]{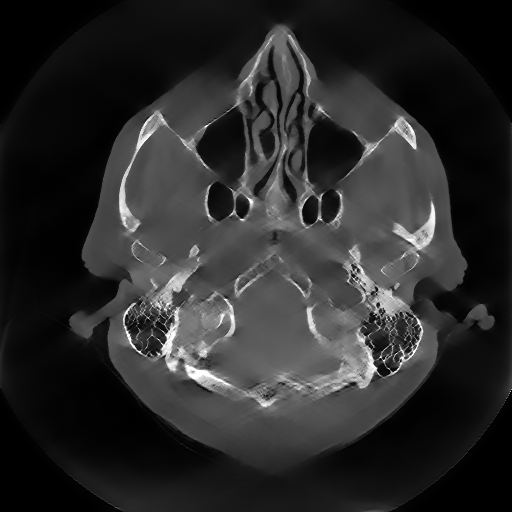}}
  }
 \end{minipage}

\caption{Reconstruction results of the $140^\circ$ (top row) and $120^\circ$ (bottom row) angular ranges at 100th iteration for wTV, ssaTV-1 ($l_{\max}=3$), and ssaTV-2 ($l_{\max}=3$), window: [-1000 1730]\,HU.}
\label{Fig:140degreeand120degree}
\end{figure}

The reconstruction results for the $140^\circ$ and $120^\circ$ angular ranges are shown in Fig.~\ref{Fig:140degreeand120degree}. For wTV, severe large streaks are observed (Figs.~\ref{Fig:140degreeand120degree}(a) and (d)). Again, they are reduced with ssaTV-1 and ssaTV-2, indicating that the proposed methods are also feasible in streak reduction for smaller angular scan ranges. However, with more data missing, several anatomical structures are degenerated.

\section{Discussion}
We observe that wTV, ssaTV-1, and ssaTV-2 reconstruct high contrast structures better than low contrast structures. This is partially because the weights in Eqn.~(\ref{eq:WeightsUpdate}) are larger for low contrast structures, which leads to a stronger TV regularization effect, blurring fine low contrast structures. 

The weight vector causes the overall minimization of $||\boldsymbol{f}||_{\text{wTV}}$ to become non-convex. The existence of local minima is a common problem in non-convex optimization. Scale-space approaches have the chance to avoid local minima since they might disappear in coarser scales and searching for the solution is more efficient than at the original scale \cite{mjolsness1991multiscale,lucia2004multi,hager2006multiscale}. That is why the proposed ssaTV algorithms reduce large streaks more efficiently and effectively than wTV. Due to the FORBILD phantom being piece-wise constant, wTV is able to reduce streaks almost as well as the two ssaTV algorithms given enough iterations in the numerical experiments. However, the clinical data has very complex structures and it suffers from more complex noise and other data inconsistencies except for Poisson noise, which potentially cause wTV to fail to reduce the large streaks here, even given more iterations.

In general, ssaTV-1 and ssaTV-2 are roughly equivalent in the sense that they both minimize the wTV term at various scales of the image, which are attained by low-pass filtering of the original image. Due to the special streak orientations in limited angle tomography, ssaTV-1 and ssaTV-2 both apply the scaling anisotropically in streaks' normal directions. Figs.~\ref{Fig:ForbildResults} and \ref{Fig:ComparisonOfAlgorithmsInClinical} demonstrate that they have a similar effect on streak reduction.


In limited angle tomography, wTV, ssaTV-1 and ssaTV-2 all suffer from missing data. For example, all three algorithms fail to exactly reconstruct the top boundaries (Figs.~\ref{Fig:ForbildResults} and \ref{Fig:ComparisonOfAlgorithmsInClinical}) and the horizontal bone structure indicated by the blue arrows (Fig.~\ref{fig:ROIclinic}). Compared with the structures in the vertical direction (Fig.~\ref{fig:ROIclinic}), those in the horizontal direction are relatively more difficult to reconstruct because no horizontal X-rays pass through the object in the $10^\circ-170^\circ$ case \cite{huang2016image}. 
\section{Conclusion}

Due to the anisotropic nature of limited angle tomography, anisotropic TV regularization is beneficial for artifact reduction. Optimization in scale space can accelerate the optimization process. In this paper, we propose two implementations of ssaTV derived from wTV with the same core idea. Both implementations apply wTV regularization in scale space and utilize low-pass filtering anisotropically to obtain coarse scale variations along streaks' normal direction. However, ssaTV-1 uses a modified gradient-like operator which considers $2\cdot s$ neighboring pixels to compute the image gradient while ssaTV-2 uses down-sampling and up-sampling operations.

Both ssaTV algorithms are investigated in numerical and clinical experiments. Compared to wTV, ssaTV-1 and ssaTV-2 reduce streak artifacts more effectively and with a high convergence speed, particularly when using multiple scaling levels. In addition, the experiments indicate that the methods are applicable   when a typical amount of noise exists.

Regarding image quality, both ssaTV-1 and ssaTV-2 reconstruct large high contrast structures very well. For medium and low contrast structures, ssaTV-1 may slightly lose spatial resolution while ssaTV-2 appears superior to wTV.

\ 

\textbf{DISCLAIMER} 

The concepts and information presented in this paper are based on research and are not commercially available.

\ifCLASSOPTIONcaptionsoff
  \newpage
\fi



%



\providecommand{\noopsort}[1]{}\providecommand{\singleletter}[1]{#1}%

%








\end{document}